\documentclass{article}

\usepackage{arxiv}

\usepackage[utf8]{inputenc} 
\usepackage[T1]{fontenc}    
\usepackage{url}            
\usepackage{booktabs}       
\usepackage{amsfonts}       
\usepackage{nicefrac}       
\usepackage{microtype}      
\usepackage{graphicx}
\RequirePackage[numbers,sort&compress]{natbib}
\usepackage{xcolor}   
\usepackage{color}
\usepackage{wrapfig}
\usepackage{siunitx}

\definecolor{cvprblue}{rgb}{0.21,0.49,0.74}
\usepackage[pagebackref,breaklinks,colorlinks,citecolor=cvprblue,hyperfootnotes=false]{hyperref}
\usepackage{doi}

\usepackage{tabularx}
\usepackage{tabulary}
\usepackage{subfig}
\newcommand{\q}[1]{``#1''}
\usepackage[accsupp]{axessibility} 
\usepackage{cleveref}       
\usepackage[para]{footmisc}

\title{ObjectDrop: Bootstrapping Counterfactuals \\ for Photorealistic Object Removal and Insertion}

\date{}

\newif\ifuniqueAffiliation

\usepackage{authblk}

\makeatletter
\renewcommand\AB@affilsepx{, \protect\Affilfont\hspace{5em}}

\newcommand\blfootnote[1]{%
  \begingroup
  \renewcommand\thefootnote{}\footnote{#1}%
  \addtocounter{footnote}{-1}%
  \endgroup
}
\makeatother

\author[1,2]{Daniel Winter}
\author[1]{\hspace{2mm}Matan Cohen}
\author[1]{\hspace{2mm}Shlomi Fruchter}
\author[1]{\hspace{2mm}Yael Pritch}
\author[1]{\hspace{2mm}Alex Rav‑Acha}
\author[1,2]{\hspace{2mm}Yedid Hoshen\vspace{5mm}}

\affil[1]{Google Research}
\affil[2]{The Hebrew University of Jerusalem}

\renewcommand{\headeright}{}
\renewcommand{\undertitle}{}
\renewcommand{\shorttitle}{ObjectDrop: Photorealistic Object Removal and Insertion}
\renewcommand{\projecturl}{https://ObjectDrop.github.io}

\hypersetup{
pdftitle={ObjectDrop: Bootstrapping Counterfactuals for Photorealistic Object Removal and Insertion},
pdfauthor={Daniel Winter, Matan Cohen, Shlomi Fruchter, Yael Pritch, Alex Rav-Acha, and Yedid Hoshen},
}

\begin{document}
\maketitle
\blfootnote{\texttt{\{daniel.winter, yedid.hoshen\}@mail.huji.ac.il}}

\begin{abstract}
 Diffusion models have revolutionized image editing but often generate images that violate physical laws, particularly the effects of objects on the scene, e.g., occlusions, shadows, and reflections. By analyzing the limitations of self-supervised approaches, we propose a practical solution centered on a \q{counterfactual} dataset. Our method involves capturing a scene before and after removing a single object, while minimizing other changes. By fine-tuning a diffusion model on this dataset, we are able to not only remove objects but also their effects on the scene. However, we find that applying this approach for photorealistic object insertion requires an impractically large dataset. To tackle this challenge, we propose bootstrap supervision; leveraging our object removal model trained on a small counterfactual dataset, we synthetically expand this dataset considerably. Our approach significantly outperforms prior methods in photorealistic object removal and insertion, particularly at modeling the effects of objects on the scene.
\end{abstract}

 \begin{figure}[!ht]
\vskip -0.5cm
    \centering
    \includegraphics[width=0.75\linewidth]{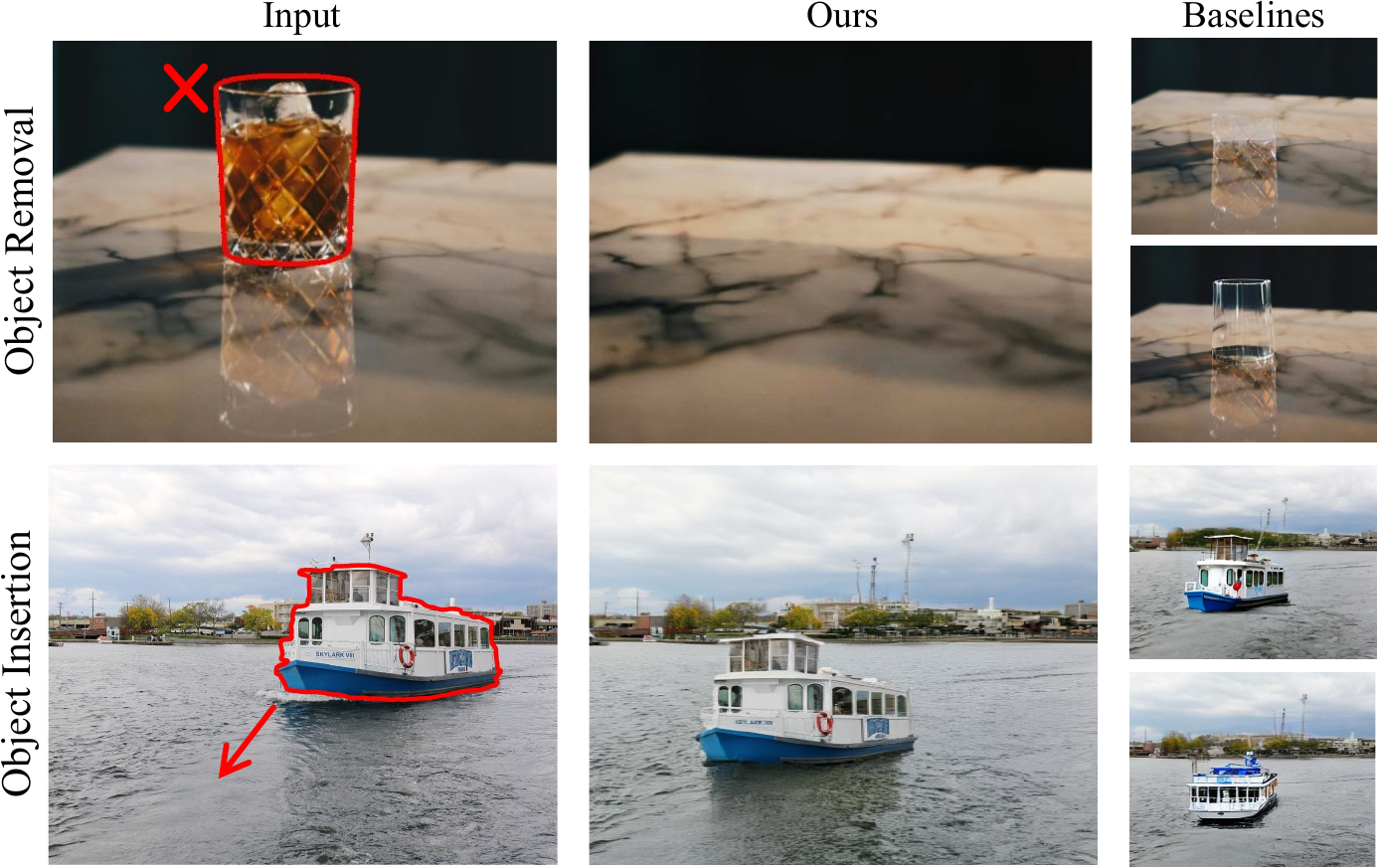}
    \caption{\textbf{Object removal and insertion.} Our method models the effect of an object on the scene including occlusions, reflections, and shadows, enabling photorealistic object removal and insertion. It significantly outperforms state-of-the-art baselines.}
    \label{fig:enter-label}
\end{figure}

\section{Introduction}
\label{sec:intro}

\begin{figure}[b]
    \vspace{-1em}
    \centering
    \includegraphics[width=0.7\linewidth]{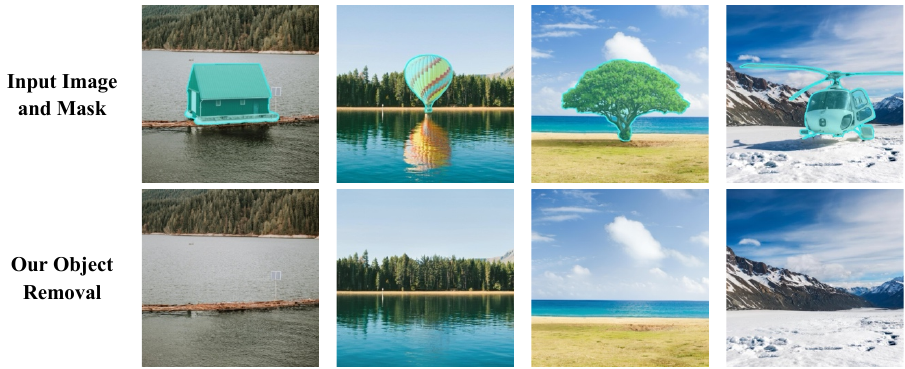}
    \caption{\textbf{Generalization.} Our counterfactual dataset is relatively small and was captured in controlled settings, yet the model generalizes exceptionally well to out-of-distribution scenarios such as removing buildings and large objects.}
    \label{fig:generalization}
\end{figure}

Photorealistic image editing requires both visual appeal and physical plausibility. While diffusion-based editing models have significantly enhanced aesthetic quality, they often fail to generate physically realistic images. For instance, object removal methods must not only replace pixels occluded by the object but also model how the object affected the scene e.g., removing shadows and reflections. Current diffusion methods frequently struggle with this, highlighting the need for better modeling of the effects of objects on their scene.

Object removal and insertion is a long-standing but challenging task. Classical image editing methods were unable to tackle the full task and instead targeted specific aspects e.g., removing hard shadows. The advent of text-to-image diffusion models enabled a new class of image editing techniques that aim to perform more general edits.

We analyze the limitations of self-supervised editing approaches through the lens of counterfactual inference. A counterfactual statement \cite{Lewis1973} takes the form "if the object did not exist, this reflection would not occur". Accurately adding or removing the effect of an object on its scene requires understanding what the scene would look like with and without the object. Self-supervised approaches rely solely on observations of existing images, lacking access to counterfactual images. Disentanglement research \cite{hyvarinen1999nonlinear,khemakhem2020variational,locatello2019challenging} highlights that it is difficult to identify and learn the underlying physical processes from this type of data alone, leading to incorrect edits.  This often manifests as either incomplete object removal or physically implausible changes to the scene.

Here, we propose a practical approach that trains a diffusion model on a meticulously curated "counterfactual" dataset. Each sample includes: i) a factual image depicting the scene, and ii) a counterfactual image depicting the scene after an object change (e.g., adding/removing it).  We create this dataset by physically altering the scene; a photographer captures the factual image, alters the scene (e.g., removes an object), and then captures the counterfactual image. This approach ensures that each example reflects only the scene changes related to the presence of the object instead of other nuisance factors of variation.

We find this technique highly effective for object removal, surprisingly even for large or inaccessible objects that were not seen in training. However, given its limited size, the same dataset proved insufficient for training the reverse task of modeling how a newly inserted object affects the scene. We hypothesize that object insertion, which requires synthesizing shadows and reflections rather than merely removing them, is inherently more complex. We expect this to require a dataset too large for us to collect.

To address this, we propose a two-step approach. First, we train an object removal model using a smaller counterfactual dataset.  Second, we apply the removal model on a large unlabeled image dataset to create a vast synthetic dataset. We finetune a diffusion model on this large dataset to add realistic shadows and reflections around newly inserted objects. We term this approach \textit{bootstrap supervision}. 

Our approach, ObjectDrop, achieves unprecedented results for both adding and removing the effects of objects. We show that it compares favorably to recent approaches such as Emu Edit, AnyDoor, and Paint-by-Example. Our contributions are:
\begin{enumerate}
\item An analysis of the limitations of self-supervised training for editing the effects of objects on scenes, such as shadows and reflections.
\item An effective counterfactual supervised training method for photorealistic object removal.
\item A bootstrap supervision approach to mitigate the labeling burden for object insertion.
\end{enumerate}

\subsection{Related Work}

\subsubsubsection{\textbf{Image inpainting.}} The task of inpainting missing image pixels has been widely explored in the literature. For several years, deep learning methods used generative adversarial network \cite{GAN} e.g. \cite{pathak2016context,hui2020image,liu2020rethinking,ntavelis2020aim,ren2019structureflow,zeng2019learning}. Several works use end-to-end learning methods \cite{iizuka2017globally,liu2018image,suvorov2022resolution,wu2022nuwa}. More recently, the impressive advancements of diffusion models \cite{sohl2015deep,song2020score,ramesh2022hierarchical, stable_diffusion}, have helped spur significant progress in inpainting  \cite{sdedit, lugmayr2022repaint, wang2023imagen, saharia2022palette, avrahami2022blended}. We show (Sec. \ref{sec:selfsupervised_limitations}) that despite the great progress in the field and using very powerful diffusion models, these methods are not sufficient for photorealistic object removal. 

\subsubsubsection{\textbf{Shadow removal methods}}
Another line of work focuses on the sub-task of shadow removal. In this task, the model aims to remove the shadow from an image given the shadow mask. Various methods \cite{SRspmnet,SRdhan,SRpmdnet,SRg2rshadownet,SRautoexposure,SRdcshadownet,SRemdn,SRsgshadownet,SRbmn,SRargan,hu2019mask,wang2018stacked} have been proposed for shadow removal. More recent methods \cite{shadowdiffusion,SRlatent}  used latent diffusion models. Unlike these methods that remove only shadows, our method aims to remove all effects of the object on the scene including: occlusions and reflections. Also, these methods require a shadow segmentation map \cite{SRdhan, wang2020instance}, while our method only requires an object segmentation map, which is easy to extract automatically e.g., \cite{kirillov2023segment}. OmniMatte \cite{lu2021omnimatte} aimed to recover both shadows and reflections, however it requires video whereas this paper deals with images.

\subsubsubsection{\textbf{General image editing model.}}
An alternative approach for removing objects from photos is to use a general purpose text-based image editing model \cite{brooks2023instructpix2pix,zhang2023hive,bar2022text2live,MGIE,emuedit}. For example, Emu Edit \cite{emuedit} trains a diffusion model on a large synthetic dataset to perform different editing tasks given a task embedding. MGIE \cite{MGIE} utilizes a diffusion model coupled with Multimodal Large Language Model (MLLM) \cite{llama,llava} to enhance the model's cross-modal understanding. While the breadth of the capabilities of these methods is impressive, our method outperformed them convincingly on object removal. 

\subsubsubsection{\textbf{Object Insertion.}}
Earlier methods for inserting an object into a new image used end-to-end Generative Adversarial Network (GAN) such as Pix2Pix \cite{pix2pix}, ShadowGAN \cite{shadowgan}, ARShadowGAN \cite{arshadowgan} and SGRNet \cite{SRGNet}. Recent studies used diffusion models. Paint-by-Example \cite{paintpyexample} and ObjectStitch \cite{objectstitch} insert a reference object into an image using the guidance of a image-text encoder \cite{clip}, but only preserve semantic resemblance to the inserted object. AnyDoor \cite{anydoor} used a self-supervised representation \cite{dino} of the reference object alongside its high-frequency map as conditions to enhance object identity preservation. While the fidelity of generated images produced by AnyDoor improved over former methods, it sometimes changes object identity entirely while we keep it unchanged by design. Furthermore, previous methods often do not model object reflections and shadows accurately, leading to unrealistic outcomes.

\begin{figure}[t]
    \centering
    \includegraphics[width=1\linewidth]{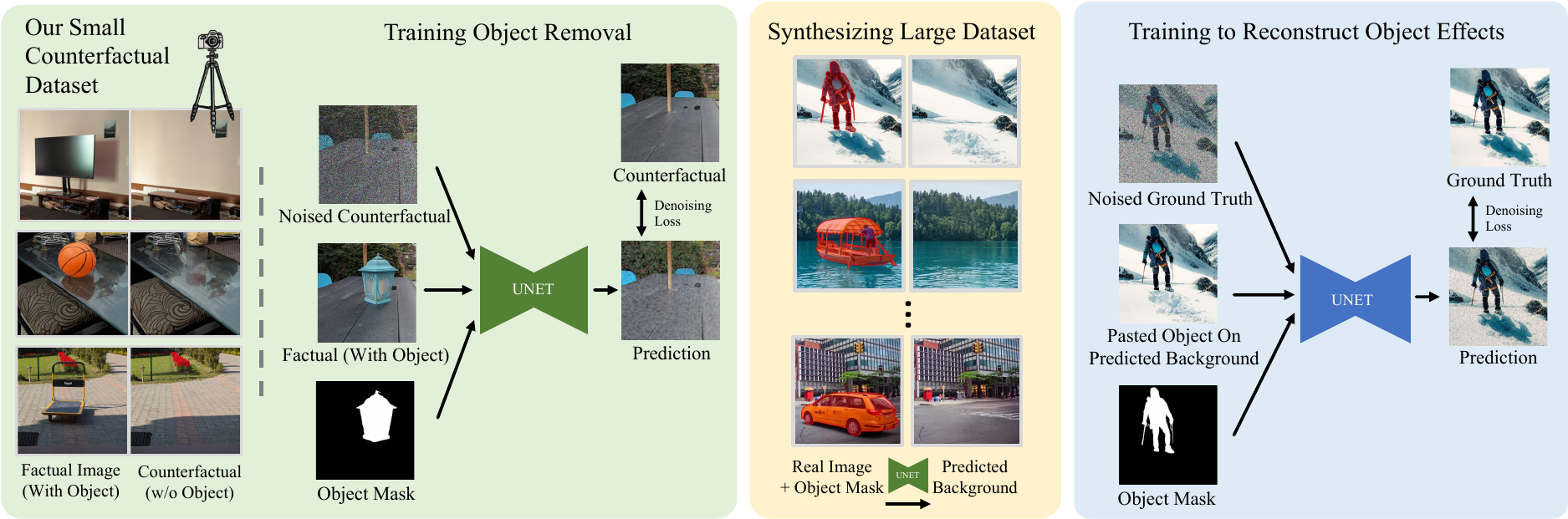}
    \caption{\textbf{Overview of our method.} We collect a \textit{counterfactual} dataset consisting of photos of scenes before and after removing an object, while keeping everything else fixed. We used this dataset to fine-tune a diffusion model to remove an object and all its effects from the scene. For the task of object insertion, we \textit{bootstrap} bigger dataset by removing selected objects from a large unsupervised image dataset, resulting in a vast, synthetic counterfactual dataset. Training on this synthetic dataset and then fine tuning on a smaller, original, supervised dataset yields a high quality object insertion model.}
    \label{fig:overview}
\end{figure}

\section{Task Definition}
\label{sec:task definition}

\begin{figure}[t]
    \centering
    \includegraphics[width=1\linewidth]{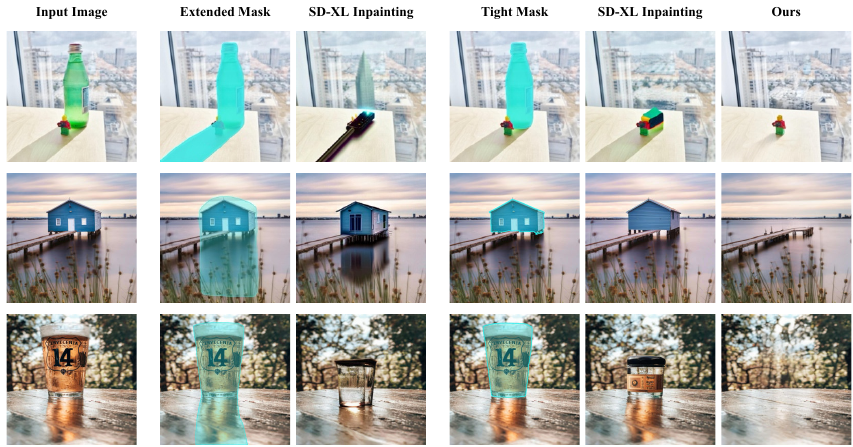}
    \caption{\textbf{Object removal - comparison with inpainting.} Our model successfully removes the masked object, while the baseline inpainting model replaces it with a different one. Using a mask that covers the reflections (extended mask) may obscure important details from the model.}
    \label{fig:compare_to_sd}
\end{figure}

\begin{figure}[t]
    \centering
    \includegraphics[width=1\linewidth]{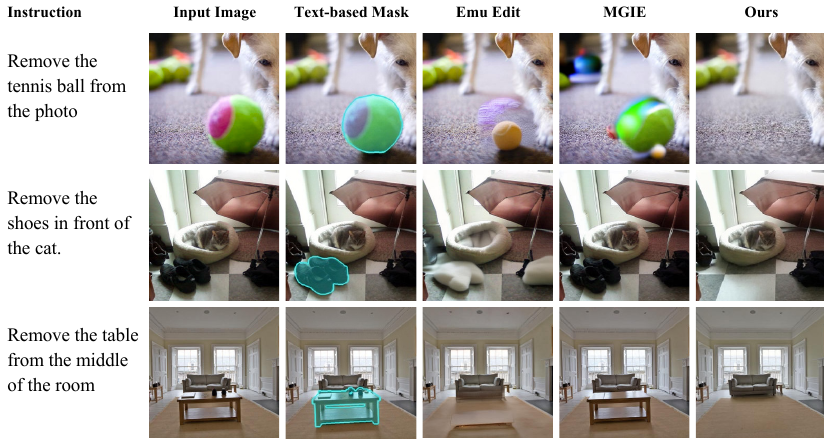}
    \caption{\textbf{Object removal - comparison with general editing methods.} We compare to general editing methods: Emu Edit and MGIE. These methods often replace the object with a new one and introduce unintended changes to the input image. For this comparison we used a text-based segmentation model to mask the object according to the instruction and passed the mask as input to our model. }
    \label{fig:compare_general_edit}
\end{figure}

We consider the input image $X$ depicting a physical $3$D scene $S$. We want our model to generate how the scene would have looked, had an object $O$ been added to or removed from it. We denote this, the \textit{counterfactual} outcome. In other words, the task is defined as re-rendering the counterfactual image $X^{cf}$, given the physical change. The challenge is to model the effects of the object change on the scene such as occlusions, shadows and reflections.
Formally, assume the physical rendering mechanism is denoted as $G_{physics}$, the input image $X$ is given by:
\begin{equation}
\label{eq:physics}
    X = G_{physics}(O, S)
\end{equation}

The desired output is the counterfactual image $X^{cf}$ s.t.,
\begin{equation}
\label{eq:physics_cf}
    X^{cf} = G_{physics}(O^{cf}, S)
\end{equation}
For object removal, an object $o$ is originally present $O=o$ and we wish to remove the object so that  $O^{cf}=\phi$ ($\phi$ is the empty object). For object insertion, the object is initially absent $O=\phi$ and we wish to add it $O^{cf}=o$. 

While the physical rendering mechanism $G_{physics}$ is relatively well understood, the formulation in Eq.~\ref{eq:physics} cannot be used directly for editing as it requires perfect knowledge of the physical object and scene, which is rarely available.

\begin{figure}[t]
\vskip -1em
    \centering
    \includegraphics[width=1\linewidth]{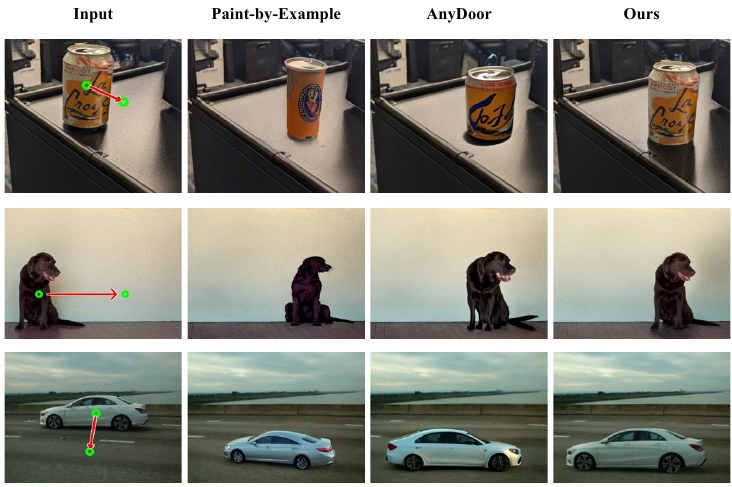}
    \caption{\textbf{Intra-image object insertion - baseline comparison.} We preserve the object identity better and achieve more photorealistic shadows and reflections than baselines Paint-by-Example and AnyDoor.}
    \label{fig:object_t2m_compare}
\end{figure}

\section{Self-Supervision is Not Enough}
\label{sec:selfsupervised_limitations}

As physical simulations are not a feasible way for photorealistic object removal and insertion in existing images, recent approaches used diffusion models instead. Diffusion models provide high-quality generative models of images i.e., they provide an effective way to estimate and sample from the distribution $P(X)$ where $X$ is a natural image. Generative models are not a natural fit for adding or removing objects from an image as they do not provide a direct way to access or modify its hidden causal variables: the physical object $O$ and the properties of the scene $S$. The task of inferring the hidden variables (e.g. $O$ and $S$) and the generative mechanism (e.g., $G_{physics}$) is called disentanglement. Several influential works \cite{hyvarinen1999nonlinear,khemakhem2020variational,locatello2019challenging} established that unsupervised disentanglement from observational data is generally impossible without strong priors. Self-supervised methods attempt to perform disentanglement using heuristic schemes. 

One common heuristic for object removal is to use diffusion-based inpainting. Such methods rely on an earlier segmentation step that splits the image pixels into two non-overlapping subsets, a subset of pixels that contain the object $X_o$ and a subset of those that do not $X_s$ s.t. $X = X_o \cup X_s$. Inpainting then uses the generative model to resample the values of the object pixels given the scene:
\begin{equation}
\label{eq:inpainting}
    x_o \sim P(X_o|X_s = x_s)
\end{equation}
The main limitation of this approach is its dependence on the segmentation mask. If the mask is chosen too tightly around the object, then $X_s$ includes the shadows and reflections of the object and thus has information about the object. The most likely values of $P(X_o|x_s)$ will contain an object that renders similar shadows and reflections as the original, which is likely a similar object to the original. If the mask is chosen so conservatively as to remove all scene pixels that are affected by the object, it will not preserve the original scene. We show both failure modes in Fig.~\ref{fig:compare_to_sd}.

Attention-based methods, such as prompt-to-prompt \cite{hertz2022prompt}, use a sophisticated heuristic based on cross-attention which sometimes overcomes the failure modes of inpainting. However, as they bias the generative model $P(X_o|X_e)$, they can result in unrealistic edits, sometimes removing the object but not its shadows. Also, the attention masks often fail to capture all scene pixels affected by the object, resulting in similar failures as inpainting. Note that Emu Edit \cite{emuedit} uses synthetic data created by an attention-based method for object removal and can therefore suffer from similar failure modes.

The core limitation of the above self-supervised approaches is the inability to directly infer the true generative mechanism and the causal hidden variables, the object $O$ and scene $S$. While heuristic methods for doing so made progress, the core limitations are hard to overcome. Class-guided disentanglement methods \cite{choi2020stargan,gabbay2021image} attempt to solve the disentanglement task from observational data by assuming perfect knowledge of one of the hidden variables (here, the physical object $O$), and assuming that the object and scene are independent. Both assumptions are not sound in this setting, as the properties of the physical object and scene are not known perfectly, and only some objects are likely in a particular scene. Note that the generative mechanism is not perfectly identifiable even when the assumptions are satisfied \cite{khemakhem2020variational,kahana2022contrastive}. This motivates our search for a more grounded approach as will be described in the following sections.

\section{Object Removal}\label{sec:removal_method}

In this section we propose ObjectDrop, a new approach based on counterfactual supervision for object removal. As mentioned in Sec. ~\ref{sec:selfsupervised_limitations}, it is insufficient to merely model the observed images, but we must also take into account their causal hidden variables i.e., the object and the scene. As we cannot learn these purely from observing images, we propose to directly act in the physical world to estimate the counterfactual effect of removing objects. 

\subsection{Collecting a counterfactual dataset.} The key for unlocking such models is by creating a counterfactual dataset. The procedure consists of three steps:
\begin{enumerate}
    \item Capture an image $X$ ("factual") containing the object $O$ in scene $S$.
    \item Physically remove the object $O$ while avoiding camera movement, lighting changes or motion of other objects.
    \item Capture another image $X^{cf}$ ("counterfactual") of the same scene but without the object $O$.
\end{enumerate}

We use an off-the-shelf segmentation model \cite{kirillov2023segment} to create a segmentation map $M_o$ for the object $O$ removed from the factual image $X$. The final dataset contain input pairs of factual image and binary object mask $(X,M_o(X))$, and the output counterfactual image $X^{cf}$.

In practice, we collected $2,500$ such counterfactual pairs. This number is relatively small due to the high cost of data collection. The images were collected by professional photographers with a tripod mounted camera, to keep the camera pose as stable as possible. As the counterfactual pairs have (almost) exactly the same camera pose, lighting and background objects, the only difference between the factual and counterfactual images is the removal of the object. 

\subsection{Counterfactual distribution estimation.}
Given our high-quality counterfactual dataset, our goal is to estimate the distribution of the counterfactual images $P(X^{cf}|X=x,M_o(x))$, given the factual image $x$ and segmentation mask. We do it by fine-tuning a large-scale diffusion model on our counterfactual dataset. We investigate the impact of using different foundational diffusion models in Sec.~\ref{sec:experiments}. The estimation is done by minimizing:

\begin{equation}
    \mathcal{L}(\theta) = \mathop{\mathbb{E}}_{t \sim U([0,T]), \epsilon \sim \mathcal{N}(0,I)}\left[\sum_{i=1}^N\|D_{\theta}(\alpha_t x^{cf}_i + \sigma_t \epsilon, x_i, M_o(x_i) , t, p) - \epsilon \|^2\right]
\end{equation}\label{eq:denoise_loss}

where $D_{\theta}(\tilde{x_t},x_{cond}, m, t, p)$ is a denoiser network with following inputs: noised latent representation of the counterfactual image $\tilde{x_t}$, latent representation of the image containing the object we want to remove $x_{cond}$, mask $m$ indicating the object's location, timestamp $t$ and encoding of an empty string (text prompt) $p$. Here, $x_t$ is calculated based on the forward process equation:

\begin{equation}
\tilde{x_t}=\alpha_t \cdot x + \sigma_t \cdot\epsilon
\end{equation}

\noindent Where $x$ represents the image without the object (the counterfactual), $\alpha_t$ and $\sigma_t$ are determined by the noising schedule, and $\epsilon \sim \mathcal{N}(0,I)$.

Importantly, unlike traditional inpainting methods, we avoid replacing the pixels of the object with uniform gray or black pixels. This approach allows our model to leverage information preserved within the mask, which is particularly beneficial in scenarios involving partially transparent objects or imperfect masks.

\begin{figure}[t]
    \centering
    \includegraphics[width=1\linewidth]{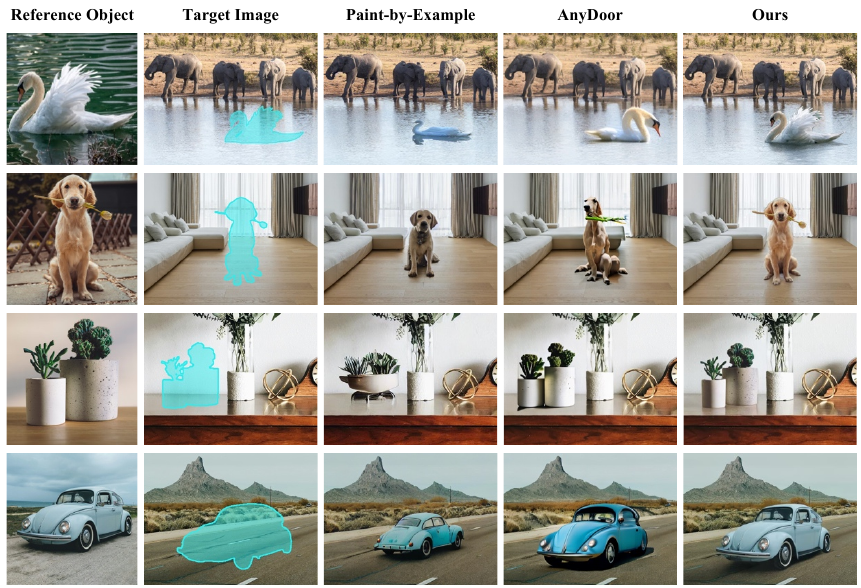}
    \caption{\textbf{Cross-image object insertion.} Similarly to the results of intra-image object insertion, our method preserves object identity better and synthesizes more photorealistic shadows and reflections than the baselines. }
    \label{fig:object_insert_compare}
\end{figure}

\subsection{Advantages over video supervision.} Our procedure requires making physical world changes, which both limits the possible size of the dataset and the scope of removed objects as they cannot be very big, very heavy or very far. It is tempting to replace this expensive supervision by using the cheaper supervision obtained from videos, as done by several previous methods including AnyDoor \cite{anydoor} and \cite{puttingpeople}. At training time, the objective of these methods is to reconstruct a masked object in one video frame, by observing the object in another frame. While this procedure is cheaper, it has serious limitations: i) In a counterfactual dataset, the only change between the images within a pair should be the removal of the object. Conversely, in video many other attributes also change, such as camera view point. This leads to spurious correlations between object removal and other attributes. ii) This procedure only works for dynamic objects (cars, animals, etc.) and cannot collect samples for inanimate objects. We show in Sec.~\ref{sec:experiments} that our method works \textit{exceptionally well}, and particularly outperforms methods that use video supervision. Furthermore, our method generalizes surprisingly well to objects that were too challenging to move in our counterfactual dataset including very heavy, large or immobile objects.

\section{Object Insertion}\label{sec:blending_method}

\begin{figure}[t]
\vskip -1em
    \centering
    \includegraphics[width=1\linewidth]{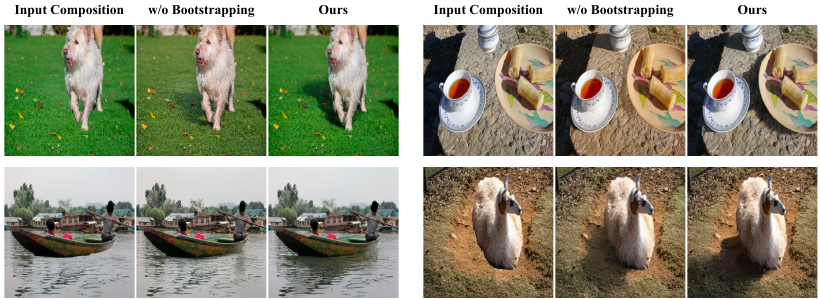}
    \caption{\textbf{Bootstrapping ablation.} Bootstrap supervision improves model quality.}
    \label{fig:ablate_bootstrap}
\end{figure}

\begin{table}[t]
\setlength\tabcolsep{10pt}
    \centering
    \caption{\textbf{Object insertion - reconstruction metrics.} A comparison with baselines: Paint-by-Example and AnyDoor, on the held-out test set. Furthermore, we ablate the contribution of the bootstrap supervision.}
    \begin{tabular}{lcccc}
        \toprule
        Model& PSNR $\uparrow$  & DINO $\uparrow$ & CLIP $\uparrow$ & LPIPS $\downarrow$ \\
        \midrule
        Paint-by-Example& 17.523& 0.755& 0.862& 0.138\\
        AnyDoor& 19.500& 0.889& 0.890& 0.095\\
        \midrule
        Ours w/o Bootstrap& 20.178& 0.929& 0.945& 0.066\\
         Ours& \textbf{21.625}&\textbf{0.939}& \textbf{0.950}&\textbf{0.057}\\
         \bottomrule
    \end{tabular}
    \label{tab:quant:insert}
\end{table}

We extend ObjectDrop to object insertion. In this task, we are given an image of an object, a desired position, and a target image. The objective is to predict how the target image would look like, had it been photographed with the given object.
While collecting a relatively small-scale (2,500 samples) counterfactual dataset was successful for object removal, we observed that this dataset is insufficient for training an object insertion model (see Fig. \ref{fig:ablate_bootstrap}). We hypothesize that this requires more examples, as synthesizing the shadows and reflections of the object may be more challenging than removing them.

\subsection{Bootstrapping counterfactual dataset.} We propose to leverage our small counterfactual dataset towards creating a large-scale counterfactual object insertion dataset. We take a large external dataset of images, and first detect relevant objects using a foreground detector. Let $x_1,x_2,...,x_n$ denote the original images and $M_o(x_1),M_o(x_2),...,M_o(x_n)$ denote the corresponding object masks. We use our object removal model $P(X^{cf}|X,M_o(X))$ to remove the object and its effects on the scene, denoting the results as $z_1,z_2,...,z_n$ where,
\begin{equation}
    z_i \sim P(X^{cf}|x_i,M_o(x_i))
\end{equation}
Finally, we paste each object into the object-less scenes $z_i$, resulting in images without shadows and reflections, 
\begin{equation}
    y_i = M_o(x_i) \odot x_i + (1 - M_o(x_i)) \odot z_i.
\end{equation}
The synthetic dataset consists of a set of input pairs $(y_i, M_o(x_i))$. The corresponding targets are the original images $x_i$. To clarify, both the input and output images contain the object $o_i$, but the input images do not contain the effects of the object on the scene, while the output images do. The task of the model is to generate the effects as illustrated in Fig.~\ref{fig:overview}.

In practice, we start with a dataset consisting of 14M images, we select 700k images with suitable objects. We run object removal on each image, and further filtered approximately half of them that did not have significant object effects on the scene. The final bootstrapped dataset consisted of 350K image, around 140 times bigger than the manually labeled dataset. Please see more details about the filtering process in the supplementary.

\subsection{Diffusion model training.}
We use the bootstrapped counterfactual dataset to train an object insertion  model with the diffusion objective presented in Eq. \ref{eq:denoise_loss}. In contrast to the object removal process, we use a pre-trained text-to-image model $D_{\theta}(x, t, p)$ that did not undergo inpainting pre-training. As the input mask increases input dimension, we add new channels to the input of the pre-trained text-to-image model, initializing their weights with $0$. 

\subsection{Fine-tuning on the ground truth counterfactual dataset.} The synthetic dataset is not realistic enough for training the final model, and is only used for pre-training. In the last stage, we fine-tune the model on the original ground truth counterfactual dataset that was manually collected. While this dataset is not large,  pre-training on the bootstrapped dataset is powerful enough to enable effective fine-tuning using this small ground truth dataset.

\section{Experiments}\label{sec:experiments}

\subsection{Implementation Details}

\textbf{Counterfactual dataset.} We created a counterfactual dataset of $2,500$ pairs of photos using the procedure detailed in Sec.~\ref{sec:removal_method}. Each pair contains a "factual" image of a scene and a second "counterfactual" image of exactly the same scene except that it was photographed after removing one object. We also held out $100$ counterfactual test examples, captured after the completion of the research, depicting new objects and scenes. 

\noindent\textbf{Model architecture.} We train a latent diffusion model (LDM) for the object removal task. We initialize using a pre-trained inpainting model, which takes as input a factual image, an object mask, and a noisy counterfactual image. We perform inference using default settings. We use a internal model with a similar architecture to Stable-Diffusion-XL. Unlike other inpainting models, we do not replace the removed object pixels with gray pixels.

\noindent\textbf{Quantitative metrics.} We compared the results of our method and the baselines on the held-out counterfactual test set. As this dataset has ground truth (see supplementary), we used standard reconstruction metrics: both classical (PSNR) and deep perceptual similarity metrics using: DINO \cite{dino}, CLIP \cite{clip}, and LPIPS \cite{lpips} (AlexNet) features.

\subsection{Object Removal}
\label{subsec:exp:removal}

\begin{figure}[t]
    \vskip -1em
    \centering
    \includegraphics[width=1\linewidth]{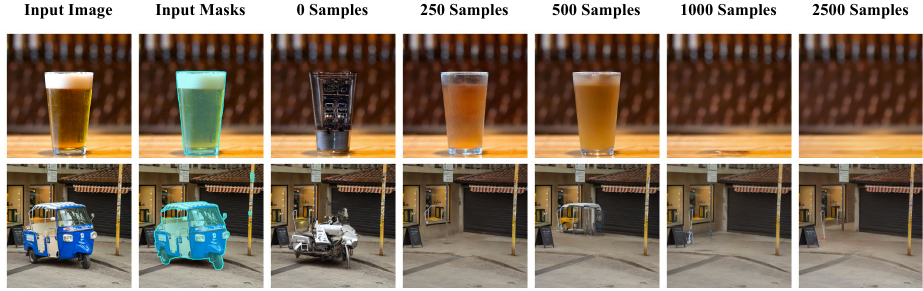}
    \caption{\textbf{Counterfactual dataset size.} Increasing the size of the training dataset improves object removal performance. The results are of high quality with $2500$ examples, but may improve further with more images.}
    \label{fig:ablate_num_samples}
\end{figure}

\begin{table}[t]
\setlength\tabcolsep{10pt}
    \centering
    \caption{\textbf{Object removal - reconstruction metrics.} A comparison with the inpainting baseline on the held-out test set.}
    \begin{tabular}{lcccc}
        \toprule
        Model& PSNR $\uparrow$ & DINO $\uparrow$ & CLIP $\uparrow$ & LPIPS $\downarrow$ \\
        \midrule
        Inpainting  & 21.192& 0.876& 0.897& 0.056\\
        Ours        & \textbf{23.153}& \textbf{0.948}& \textbf{0.959}& \textbf{0.048}\\
        \bottomrule
    \end{tabular}
    \label{tab:quant:remove}
\end{table}

\textbf{Qualitative results.} We evaluated our result on the benchmark published by Emu Edit \cite{emuedit}. As seen in Fig.~\ref{fig:compare_general_edit}, our model removes objects and their effects in a photorealistic manner. The baselines failed to remove shadows and reflections and sometimes adversely affected the image in other ways.

\noindent\textbf{Quantitative results.} Tab.~\ref{tab:quant:remove} compares our method to the inpainting pre-trained model on the held-out test set using quantitative reconstruction metrics. Our method outperformed the baseline substantially.

\noindent\textbf{User study.} We conducted a user study on the benchmark by Emu Edit \cite{emuedit} between our method and baselines: Emu Edit and MGIE. As the benchmark does not have ground truth, user study is the most viable comparison. We used the CloudResearch platform to gather user preferences from $50$ randomly selected participants. Each participant was presented with $30$ examples of an original image, removal text instructions, and results generated by our method and the baseline. Tab.~\ref{tab:userstudy:removal} displays the results. Notably, our method surpassed both baseline methods in user preference.

\begin{table}[t]
\setlength\tabcolsep{15pt}
  \caption{\textbf{Object removal - user study.} A comparison to Emu Edit and MGIE on the Emu Edit dataset \cite{emuedit}}
  \label{tab:userstudy:removal}
  \centering
  \begin{tabularx}{0.8\textwidth}{XrlXr}
    \toprule
    \multicolumn{5}{c}{Which model did better job at following the object removal editing instruction?}\\
    \midrule
    Preferred Emu Edit & 35.9\% & & Preferred MGIE   & 13.5\%\\
    Preferred ours     & 64.1\% & & Preferred ours   & 86.5\%\\
  \bottomrule
 \end{tabularx}
\end{table}

\begin{figure}[t]
\vskip -1em
    \centering
    \includegraphics[width=1\linewidth]{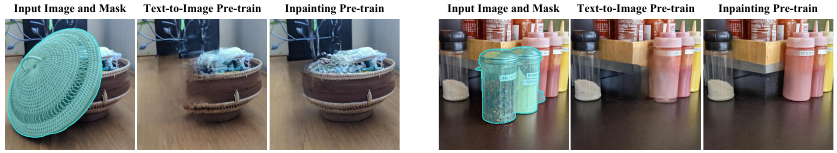}
    \caption{\textbf{Inpainting vs. Text-to-Image pre-training.} The inpainting model has better results on pixels occluded by the objects but results in comparable quality for removing or adding photorealistic reflections and shadows.}
    \label{fig:ablate_pretrain}
\end{figure}

\begin{table}[t]
\caption{\textbf{Object insertion - user study.} A comparison on in-distribution (ID) and out-of-distribution (OOD) intra-image object insertion datasets.}
\label{tab:insertion_user_study}
\centering
\setlength\tabcolsep{6pt}
\begin{tabularx}{0.48\textwidth}{lrXr} 
\toprule
\multicolumn{4}{c}{\textbf{Held-Out Set (ID)}}\\
\midrule
AnyDoor & 11.1\% &  Paint-by-Example & 3.3\% \\
Ours & \textbf{88.9\%} & Ours & \textbf{96.7\%} \\
\bottomrule
\end{tabularx}
\hfill 
\begin{tabularx}{0.48\textwidth}{lrXr} 
\toprule
\multicolumn{4}{c}{\textbf{In-the-Wild (OOD)}}\\
\midrule
AnyDoor & 5.0\% & Paint-by-Example & 2.8\% \\
Ours & \textbf{95.0\%} & Ours & \textbf{97.2\%} \\
\bottomrule
\end{tabularx}
\end{table}

\begin{table}[!ht]
\setlength\tabcolsep{10pt}
    \centering
    \caption{\textbf{Stable-Diffusion results.} Our method works well on the public Stable-Diffusion-Inpainting v1 model.}
    \label{tab:quant_sd_remove}
    \begin{tabular}{lcccc}
        \toprule
        Model& PSNR $\uparrow$  & DINO $\uparrow$ & CLIP $\uparrow$ & LPIPS $\downarrow$ \\
        \midrule
        SD Inpainting    & 19.198& 0.775& 0.884& 0.083\\
        Ours SD          & \textbf{21.363}& \textbf{0.876}& \textbf{0.930}& \textbf{0.076}\\
        \bottomrule
    \end{tabular}
\end{table}

\subsection{Object Insertion}

\textbf{Qualitative results.} We compare our object insertion model with state-of-the-art image reference-based editing techniques, Paint-by-Example \cite{paintpyexample} and AnyDoor \cite{anydoor}. Fig.~\ref{fig:object_t2m_compare} shows intra-image insertions, i.e., when the objects are re-positioned within same image. For achieving intra-image insertions we first use our object removal model to remove the object from its original position, obtaining the background image. For equitable comparisons, we used the same background image, obtained by our model, when comparing to the baselines. Fig.~\ref{fig:object_insert_compare} shows inter-image insertions, i.e., when the objects come from different images. In both cases our method synthesizes the shadows and reflections of the object better than the baselines. It also preserves the identity of the object, while other methods modify it freely and in many cases lose the original identity entirely.

\noindent\textbf{Quantitative results.} We compare to Paint-by-Example and AnyDoor on the held-out counterfactual test dataset. The results are presented in Tab.~\ref{tab:quant:insert}. Our method outperforms the baselines by a significant margin on all metrics. 

\noindent\textbf{User study.} We also conducted a user-study on 2 intra-image insertion datasets. The first is the held-out test set. The second is a set of $50$ out-of-distribution images depicting more general scenes, some are very different from those seen in training e.g., inserting boats and building. Tab.~\ref{tab:insertion_user_study} shows that users overwhelmingly preferred our method over the baselines.

\subsection{Ablation Study}

\noindent\textbf{Bootstrapping.} We ablated the contribution of our bootstrapping method for object insertion. Here, we bootstrapped $2,500$ real images into $350K$ synthetic images. In the ablation, we compare our full method to finetuning the original backbone on the original counterfactual dataset without bootstrapping. Both models used the same pretraining backbone. Both the qualitative results in Fig.~\ref{fig:ablate_bootstrap}  and the quantitative results in Tab.~\ref{tab:quant:insert} clearly support bootstrapping.

\noindent\textbf{Dataset size.} Collecting large counterfactual datasets is expensive. We evaluate the influence of dataset size on the performance of our object removal method. We finetune the base model on subsets of the full counterfactual dataset with varying sizes. Fig.~\ref{fig:ablate_num_samples} shows that using the pre-trained inpainting model without finetuning ("0 samples") merely replaces the target object by another similar one. Also, its effects on the scene remain. The results start looking attractive around $1000$ counterfactual examples, more examples further improve performance. 

\noindent\textbf{Text-to-image vs. inpainting pretrained model.} Fig.~\ref{fig:ablate_pretrain} demonstrates that using a text-to-image (T2I) instead of an inpainting model for pre-training obtains comparable quality for removing shadows and reflections. This shows that inpainting models do not have better inductive bias for modeling object effects on scenes than T2I models. Unsurprisingly, the inpainting model is better at inpainting the pixels occluded by the objects. Furthermore, we compared the pre-trained models (not shown) on object insertion. Consequently, we used the inpainting backbone for the object removal experiments and the T2I backbone for the object insertion experiments in the paper.

\noindent\textbf{Public models.} We verified that our method works on publicly available models. Here, we trained our model using Stable-Diffusion-Inpainting~v1 as the pre-trained backbone. We then computed the quantitative metrics for object removal as in Sec.~\ref{subsec:exp:removal}. Our results in Tab.~\ref{tab:quant_sd_remove} show that our method improves this pretrained model significantly.

\section{Limitations}
This work focuses on simulating the effect that an object has on the scene, but not the effect of the scene on the object. As a result, our method may yield unrealistic results in scenarios where the orientation and lighting of the object are incompatible with the scene. This can be solved independently using existing harmonization methods, but this was not explored in the context of this work. Additionally, as our model does not know the physical 3D scene and lighting perfectly, it may result in realistic-looking but incorrect shadow directions.

\section{Conclusion}
We introduced ObjectDrop, a supervised approach for object removal and insertion to overcome the limitations of previous self-supervised approaches. We collected a counterfactual dataset consisting of pairs of images before and after the physical manipulation of the object. Due to the high cost of obtaining such a dataset, we proposed a bootstrap supervision method. Finally, we showed through comprehensive evaluation that our approach outperforms the state-of-the-art.

\section{Acknowledgement}
We would like to thank to Gitartha Goswami, Soumyadip Ghosh, Reggie Ballesteros, Srimon Chatterjee, Michael Milne and James Adamson for providing the photographs that made this project possible. We thank Yaron Brodsky, Dana Berman, Amir Hertz, Moab Arar, and Oren Katzir for their invaluable feedback and discussions. We also appreciate the insights provided by Dani Lischinski and Daniel Cohen-Or, which helped improve this work.

\clearpage
\bibliographystyle{ieeenat_fullname}
\bibliography{references}

\appendix

\section{Training and Inference}

\subsubsection{Object Removal}
In our object removal training process, we utilize a pre-trained text-to-image latent diffusion model (LDM) that was further trained for inpainting. Given an image of an object ("factual") and its mask, we finetune the LDM to denoise an image of the same scene without the object (the counterfactual image). We performed $50,000$ optimization steps with batch size of $128$ images and learning rate of $1e\text{--}5$.

\subsubsection{Object Insertion}
To train our object insertion model, we first finetune the model using a synthetic dataset as described in Section 5. This initial training phase consists of $100,000$ optimization steps, employing a batch size of $512$ images and a learning rate of $5e\text{--}5$. Subsequently, we fine-tune the model on our original counterfactual dataset for an additional $40,000$ steps, with batch size of $128$ and decaying learning rates.

\noindent The denoiser function $D_{\theta}(x_t,x_{cond}, m, t, p)$ receives the following inputs:
\begin{itemize}
\item $x_{t}$: Noised latent representation of the image containing the object.
\item $x_{cond}$: Latent representation of the object pasted onto a background image as is, without its effects on the scene.
\item $m$: Mask indicating the object's location.
\item $t$: Timestamp.
\item $p$: Encoding of an empty string (text prompt).
\end{itemize}

\subsection{Inference}
All images in this paper were generated at resolution of $512\times512$, with 50 denoising steps. 

\section{Bootstrapping}
The bootstrapping procedure for creating the object insertion training set, as outlined in Section 5, follows these steps: We begin with an external dataset of $14$ million images and extract foreground segmentation for each. We filter out images where the foreground mask covers less than $5\%$ or more than $50\%$ of the total image area, aiming to exclude objects that are either too small or too large. Additionally, we eliminate images where the foreground object extends across more than $20\%$ of the lower image boundary, as the shadow or reflection of these objects is often not visible within the image. This filtering process results in $700,000$ images potentially containing suitable objects for removal. Using our object removal model, we generate predicted background images. However, in many of the original images, the object does not have a significant shadow or reflection, so that the difference between the synthesized input and output pairs consists of noise. To address this, we further discard images where the area showing significant differences between the object image and the predicted background is too small. This yields our final bootstrapped dataset of $350,000$ examples.

\section{Evaluation Datasets}\label{sec:testsets}

\begin{figure}[t]
\vspace{-1.5em}
    \centering
    \includegraphics[width=1\linewidth]{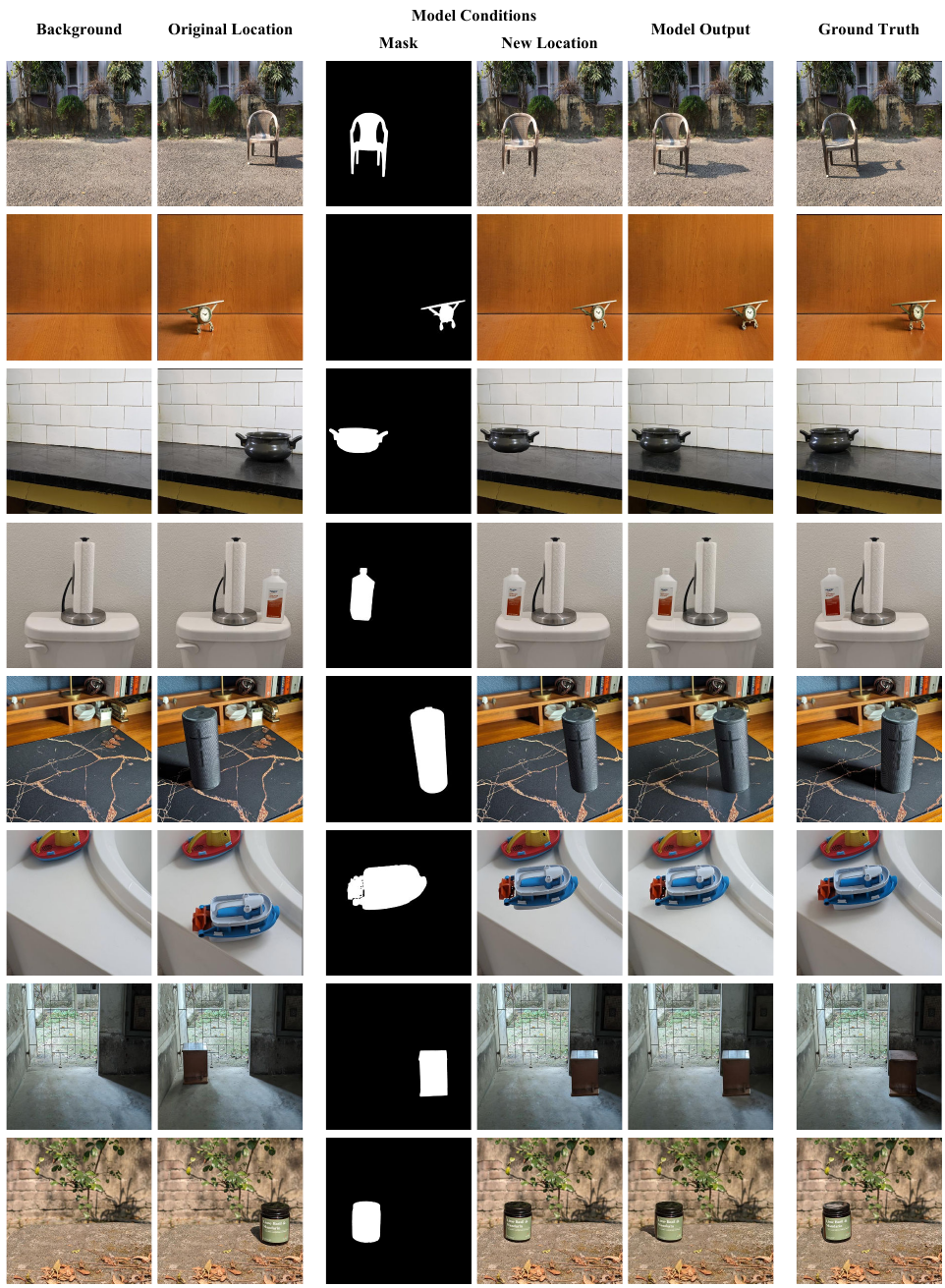}
    \caption{Our held-out test set. The object insertion model uses two conditions: (1) An image where the object was pasted naively on the background and (2) a mask of that object.}
    \label{fig:held_out}
\end{figure}

To assess our object insertion model, we employed two datasets. The first, referred to as the held-out dataset, comprises 51 triplets of photos taken after the completion of the project. Each triplet consists of: (1) a scene without the object, (2) the same scene with an added object, and (3) another image of the same scene and the same object placed elsewhere. We automatically segmented \cite{kirillov2023segment} the added object and relocated it within the image by naively pasting it on the background scene image. The model's inputs consist of the image with the pasted object and its corresponding mask. This dataset, along with our results, are presented in Fig. \ref{fig:held_out}. With ground truth images illustrating how object movement should appear, we conducted quantitative metric assessments and user studies. Additionally, we used this dataset for evaluating the object removal model. In this test, we removed the object and compared the generated image to the ground truth background image.

\begin{figure}[t]
\vspace{-2em}
    \centering
    \includegraphics[width=1\linewidth]{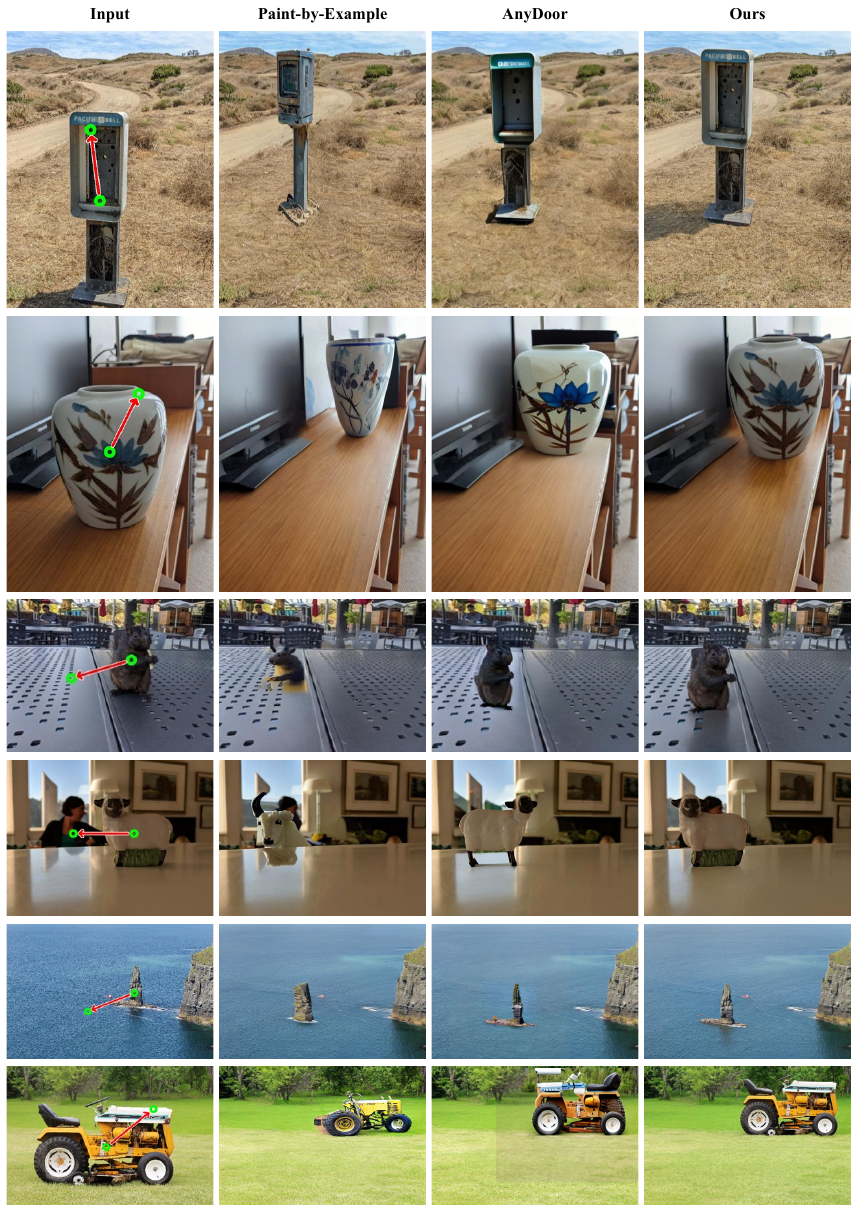}
    \caption{Additional examples of inta-image object insertion.}
    \label{fig:tap_to_move_extended}
\end{figure}

The second test set, utilized for object insertion, comprises 50 examples, including some out-of-distribution images intended for moving large objects, as shown in Fig. \ref{fig:tap_to_move_extended}. As this dataset is lacking ground truth images, we used this dataset solely for user study.

\section{User Study}
\begin{figure}[t]
    \vspace{-2em}
    \centering
    \includegraphics[width=0.98\linewidth]{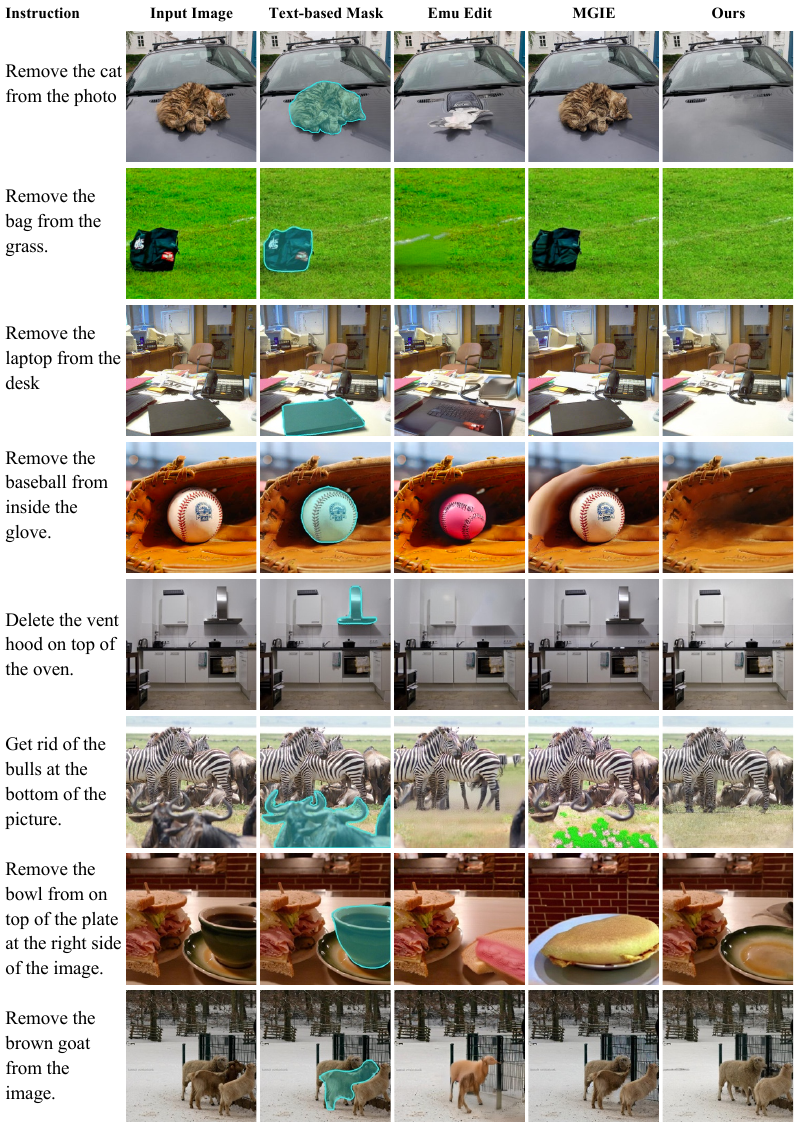}
    \caption{Additional examples for comparison with general editing methods, Emu Edit and MGIE. In this comparison, we utilized a text-based segmentation model to generate a mask for the object based on given instructions, which was then used as input for our model.}
    \label{fig:emu_testset_extended}
\end{figure}

To assess the effectiveness of our object removal model, we conducted a user study using the test set provided by Emu Edit of 264 examples, as shown in Fig. \ref{fig:emu_testset_extended}. We compared our results separately with those of Emu Edit and MGIE. Utilizing the CloudResearch platform, we collected user preferences from 50 randomly selected participants. Each participant reviewed 30 examples consisting of an original image, removal instructions, and the outcomes produced by both our method and the baseline. We randomized both the order of the examples shown and the order of each model in each example. To improve the reliability of the responses, we duplicated a few questions, and removed questionnaires that showed inconsistency for those repeated questions. A similar user study was carried out to compare our object insertion model with AnyDoor and Paint-by-Example, using the datasets described in Section \ref{sec:testsets}. Different participants were used for each dataset and comparison with baselines. The majority of participants were located in the United States. Participant were compensated above the minimum wage.

\begin{figure}[t]
\vspace{-1.5em}
    \centering
    \includegraphics[width=0.9\linewidth]{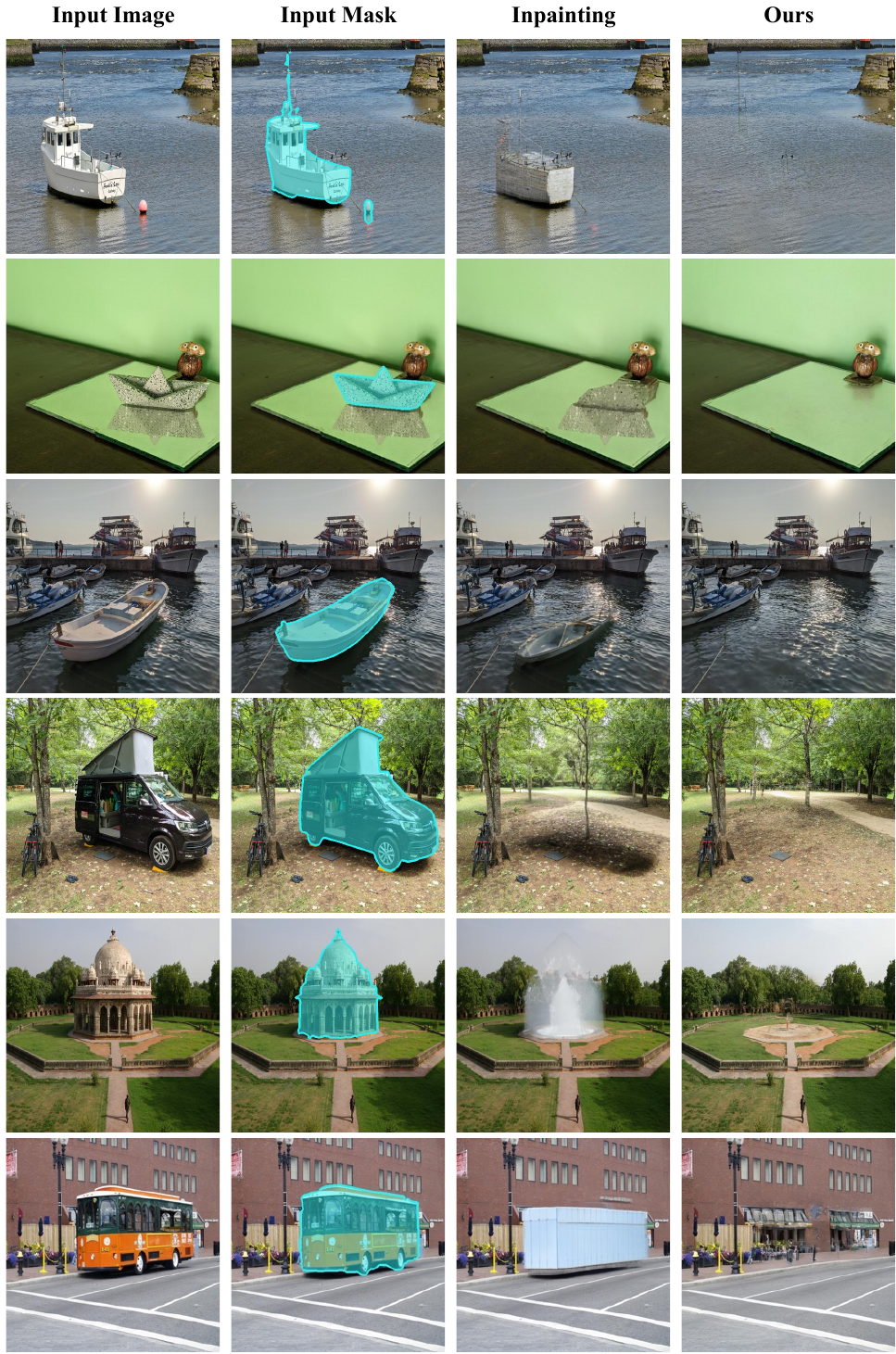}
    \caption{Additional examples showcasing the performance of our object removal model compared to the inpainting model we initialized our model with.}
\end{figure}
\begin{figure}[t]
\vspace{-1.5em}
    \centering
    \includegraphics[width=0.9\linewidth]{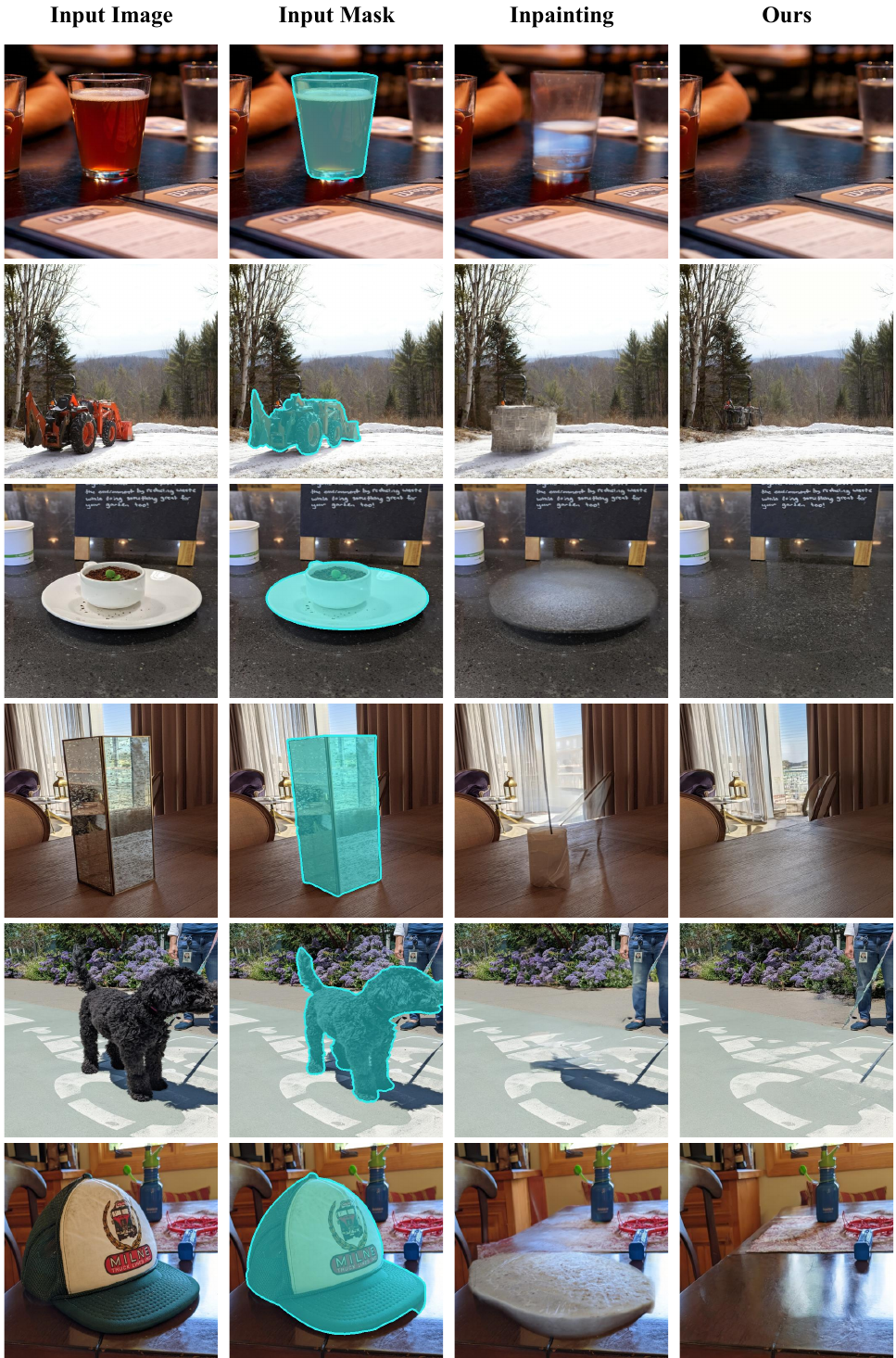}
    \caption{Additional examples showcasing the performance of our object removal model compared to the inpainting model we initialized our model with.}
\end{figure}
\begin{figure}[t]
\vspace{-1.5em}
    \centering
    \includegraphics[width=0.9\linewidth]{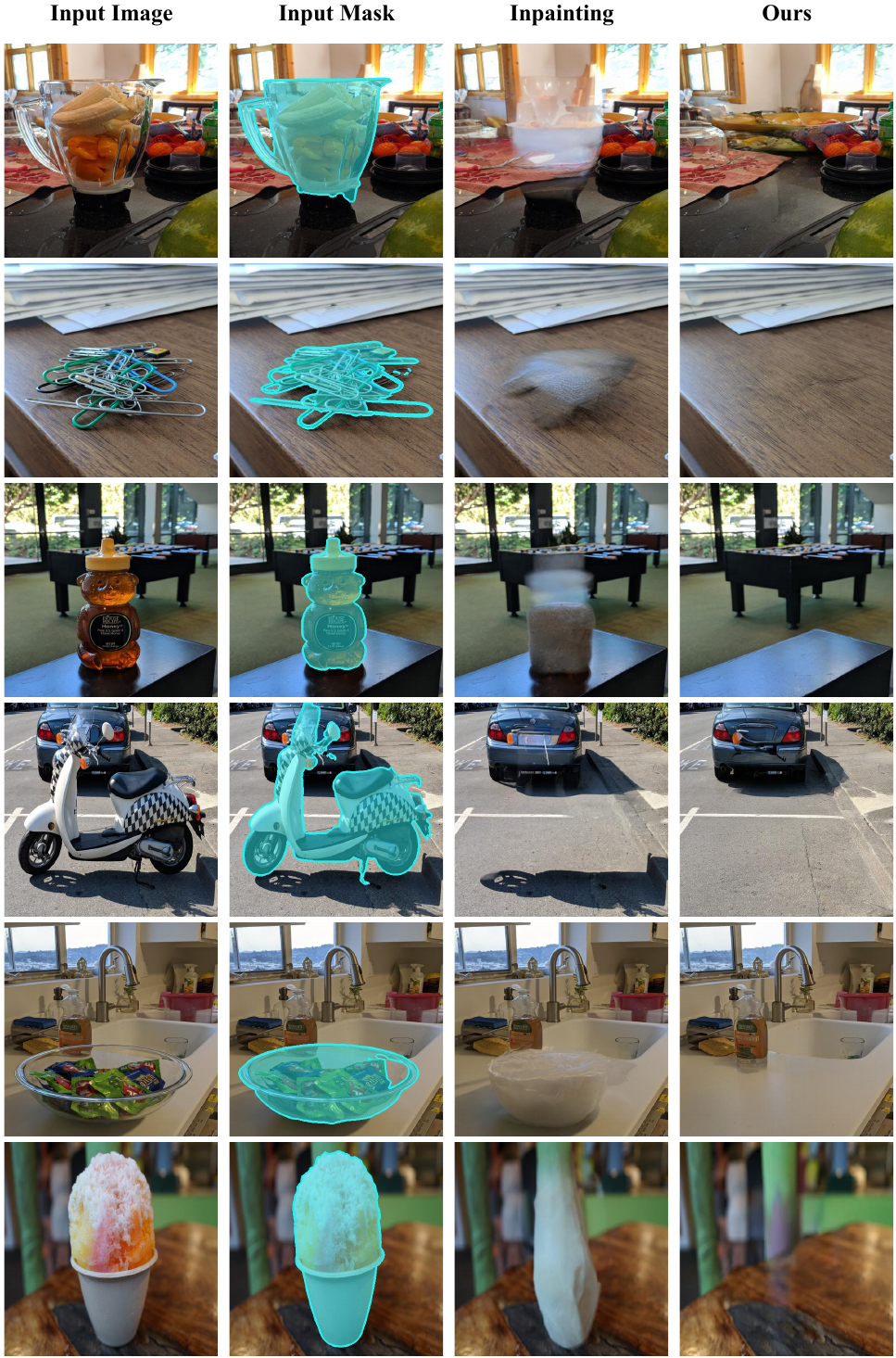}
    \caption{Additional examples showcasing the performance of our object removal model compared to the inpainting model we initialized our model with.}
\end{figure}

\end{document}